%% file: main-arxiv.tex
\documentclass{article}
\usepackage{authblk}
\usepackage{xcolor}
\usepackage[
  left=1in,
  right=1in,
]{geometry}
\usepackage{hyperref}
\usepackage{natbib}
\usepackage{graphicx}
\usepackage{booktabs}
\usepackage{subcaption}

\input{our-macros}

\usepackage{our_style}

\title{Context-Informed Ship Trajectory Prediction via Conditional Attention}
\author[1]{Yuan Guan}
\author[1]{Chandler Squires}
\author[2]{Timothy Hu}
\author[1]{Pradeep Ravikumar}
\affil[1]{Machine Learning Department, Carnegie Mellon University}
\affil[2]{Advanced Technology Laboratories
Lockheed Martin}
\date{}

\begin{document}

\maketitle

\let\thefootnote\relax\footnotetext{\copyright~2026 IEEE. Personal use of this material is permitted. Permission from IEEE must be obtained for all other uses, in any current or future media, including reprinting/republishing this material for advertising or promotional purposes, creating new collective works, for resale or redistribution to servers or lists, or reuse of any copyrighted component of this work in other works.}

\begin{abstract}
Long-term ship trajectory prediction is a fundamental capability for maritime safety and autonomous navigation. While recent Transformer-based architectures have improved forecasting horizons, they predominantly rely on historical kinematic states, treating vessel motion as an isolated system. In reality, maritime navigation is profoundly modulated by extrinsic factors like weather and constrained by static vessel characteristics. Existing multimodal approaches fundamentally model the joint distribution over states and contexts, treating environmental variables as peer features rather than encoding the directional physical dependence of vessel dynamics on environmental conditions.

In this work, we propose the Conditional Informer, a novel encoder-decoder architecture that formulates trajectory prediction as a conditional generation task. We employ a dedicated Conditional Attention mechanism where the vessel state explicitly queries environmental contexts through cross-attention, encoding the physical prior that weather modulates—but is not generated by—vessel dynamics. Furthermore, to address the intermittency of real-world data, we introduce a Modality Masking training strategy to prevent catastrophic degradation during sensor fallback. Extensive experiments on AIS and ERA5 data demonstrate that our approach outperforms kinematic and concatenation-based baselines by 15.4\% in prediction accuracy when context is available. Crucially, Modality Masking prevents shortcut learning, reducing fallback error by nearly an order of magnitude compared to unconstrained models.
\end{abstract}

\input{content/intro}
\input{content/setup}
\input{content/methods}
\input{content/experiments}
\input{content/results}
\input{content/discussion}

\paragraph{Acknowledgements}
This research was developed with funding from the Defense Advanced Research Projects Agency (DARPA) via HR0011-25-3-0239, FA8750-23-2-1015, ONR via N00014-23-1-2368, and NSF via IIS-1909816.
\bibliographystyle{plainnat}
\bibliography{bib}
\end{document}

%% file: our-macros.tex
\definecolor{darkgreen}{rgb}{0,0.5,0}

\hypersetup{
    unicode=false,
    colorlinks=true,
    linkcolor=blue,
    citecolor=darkgreen,
    urlcolor=magenta
}

\newcommand{\defeq}{\mathrel{:=}}

\newcommand{\bbN}{\mathbb{N}}
\newcommand{\bbR}{\mathbb{R}}

\newcommand{\ba}{\mathbf{a}}
\newcommand{\bw}{\mathbf{w}}
\newcommand{\bx}{\mathbf{x}}

\newcommand{\bW}{\mathbf{W}}
\newcommand{\bX}{\mathbf{X}}

\newcommand{\cA}{\mathcal{A}}
\newcommand{\cW}{\mathcal{W}}
\newcommand{\cX}{\mathcal{X}}

\newcommand{\metadim}{{d_a}}
\newcommand{\statedim}{{d_x}}
\newcommand{\contextdim}{{d_w}}

\newcommand{\past}{{\text{past}}}
\newcommand{\None}{\textsc{None}}

%% file: content/intro.tex
\section{Introduction}

As autonomous vessel technologies advance, the ability to forecast a vessel's future state---position, speed, and course---over long horizons becomes increasingly critical. 
While the Automatic Identification System (AIS) is the primary source
of real-time vessel telemetry, its signals suffer dropouts from transmission failures, antenna obstructions, satellite coverage
gaps, or deliberate transponder deactivation. In such scenarios,
accurate trajectory prediction models are essential for maintaining
maritime domain awareness, enabling operators to track vessels
through AIS gaps and to deconflict shipping lanes proactively.
Beyond AIS, similar kinematic features can be derived through
satellite imagery and radar-based tracking~\cite{milios2019automatic},
though at lower temporal resolution and higher computational cost.
Deep learning models, particularly Transformers, have largely superseded RNNs~\cite{capobianco2021deep} and kinematic models (e.g., Kalman Filters) for AIS trajectory forecasting by effectively capturing long-range dependencies. Notably, Informer-TP~\cite{xiong2024informer} utilizes ProbSparse Attention to significantly reduce long-term error accumulation compared to standard Transformers. 

Despite these advances, a critical limitation remains: state-of-the-art models treat vessel motion as an isolated kinematic process, relying exclusively on historical trajectory states. In reality, a ship is an open physical system whose trajectory is profoundly modulated by extrinsic environmental forces (such as winds, waves, and currents) and constrained by static vessel characteristics. While a broad range of exogenous factors---including seasonal shipping patterns, port congestion, and geopolitical events~\cite{marinetraffic}---may influence routing decisions, we focus on the physically measurable environmental forces (wind speed and significant wave height) that exert direct and continuous mechanical influence on vessel dynamics and are available globally through reanalysis products such as ERA5. The conditional architecture we propose is, however, extensible to additional exogenous signals as they become available.

In this work, we propose the Conditional Informer, a novel architecture that extends the Informer-TP framework to explicitly model the conditional dependence of trajectory states on environmental context and static metadata. Our approach utilizes separate embedding pathways and a specialized Conditional Attention mechanism to capture the effect of these other variables on the evolution of ship states.

\input{figures/data_img}

\subsection{Multimodal Data Fusion}

Recognizing the importance of external factors, recent research has begun to explore Multimodal Trajectory Prediction. The core challenge lies in effectively fusing high-frequency, sparse kinematic data (AIS) with low-frequency, dense environmental fields (weather) and static attributes (metadata). Existing approaches generally fall into two categories:
\begin{enumerate}
    \item \textbf{Parallel Feature Extraction:} \citet{zhang2024adaptive} proposed the Adaptive Multimodal Data (AMD) model, which employs separate feature extraction branches using GRUs for AIS and MLPs for environmental data before merging them in a central fusion block. 
    \item \textbf{Gated Fusion:} 
    \citet{chen2023vessel} introduced the Dual-Path Spatial-Temporal Attention Network (DualSTMA), which injects static vessel attributes into the dynamic stream of the Transform via a gating mechanism that modulates the learned features. 
\end{enumerate}  
While these architectures move beyond naïve input concatenation, they model the joint distribution $P(X,W)$ over states ($X$) and contexts ($W$), treating environmental variables as peer features that help explain variance in the output. Critically, this symmetric treatment does not encode the directional physical relationship in the maritime system: environmental forces influence vessel dynamics, but the vessel's motion does not alter the weather. As we demonstrate empirically in Section V-D, this omission has practical consequences—without structural constraints enforcing the correct directionality, multimodal models are susceptible to shortcut learning~\citep{geirhos2020shortcut}, collapsing onto the strongest available signal while suppressing essential but weaker modalities.

\subsection{Causal Perspective}

Environmental conditions and vessel dynamics exhibit a strict physical asymmetry: weather perturbs vessel motion, but not vice versa.

 Following~\citet{scholkopf2021toward}, encoding this known causal structure as an architectural inductive bias---separating the environment encoder from the state pathway---produces models that generalize more robustly than joint representations.

The practical risk of ignoring this asymmetry is \textit{shortcut learning}~\citep{geirhos2020shortcut}: when multiple input modalities carry different levels of predictive signal, unconstrained models can exploit the strongest signal while suppressing weaker but essential features. In multimodal trajectory prediction, this manifests as \textit{feature collapse}---the model learns to rely entirely on environmental context and disregards kinematic state, leading to catastrophic failure when context is unavailable at inference.

We therefore adopt the design principle that the known physical hierarchy should be reflected in the model architecture. In our Conditional Attention mechanism, the vessel state generates queries while the environment provides keys and values, forming directional attention matrices where the state attends to the context but not vice versa. %This directional structure is necessary to encode the physical asymmetry, but not sufficient to prevent shortcut learning on its own---Modality Masking (Section~\ref{sec:modality_dropout}) complements it by forcing the model to maintain a viable state-only prediction pathway throughout training.

%% file: figures/data_img.tex
\begin{figure}[tbhp]
\centering
% Reduce column padding globally for this figure to fit all tables natively
\setlength{\tabcolsep}{3.5pt} 

\begin{subfigure}[b]{0.47\textwidth}
\centering
\begin{tabular}{c | c c c c}
\toprule
$t$ & LAT (°N) & LON (°W) & SOG (kn) & COG (°) \\
\midrule
1 & 23.8 & -83.5 & 14.2 & 072 \\
2 & 24.9 & -81.6 & 14.8 & 075 \\
3 & 26.3 & -79.8 & 15.1 & 078 \\
\vdots & \vdots & \vdots & \vdots & \vdots \\
\bottomrule
\end{tabular}
\caption{State Trajectory $\bX^k$}
\end{subfigure}
\hfill
\begin{subfigure}[b]{0.25\textwidth}
\centering
\begin{tabular}{c | c c}
\toprule
$t$ & SWH ($m$) & WS ($ms^{-1}$) \\
\midrule
1 & 0.95 & 8.59 \\
2 & 0.99 & 7.52 \\
3 & 0.99 & 6.49 \\
\vdots & \vdots & \vdots \\
\bottomrule
\end{tabular}
\caption{Context Trajectory $\bW^k$}
\end{subfigure}
\hfill
\begin{subfigure}[b]{0.26\textwidth}
\centering
\begin{tabular}{c | c}
\toprule
$t$ & DateTime (UTC) \\
\midrule
1 & 2023-06-30 14:31:00 \\
2 & 2023-06-30 14:36:00 \\
3 & 2023-06-30 14:41:00 \\
\vdots & \vdots \\
\bottomrule
\end{tabular}
\caption{Timestamps}
\end{subfigure}

\caption{
\textbf{Example state and context trajectories for context-informed ship trajectory prediction.}
In (a), LAT, LON, SOG, and COG are short for \emph{latitude}, \emph{longitude}, \emph{speed over ground}, and \emph{course over ground}, respectively.
In (b), SWH and WS are short for \emph{significant wave height} and \emph{wind speed}, respectively.
}
\label{fig:sample-tabular-data}
\end{figure}

%% file: content/setup.tex
\section{Problem Formulation}

In general, we introduce the task of \emph{context-informed trajectory prediction}.
Let $k$ be an index for the targets whose trajectories we wish to predict.
Each index $k$ is associated with a \emph{trajectory length} $T_k \in \bbN$ and \emph{state trajectory} $\bX^k$.
Further, each $k$ may optionally be associated with a \emph{context trajectory} $\bW^k$, and a \emph{metadata vector} $\ba^k$, as follows.

\subsection{States, Contexts, and Metadata}

Letting $\cX = \bbR^\statedim$ denote the \emph{state space} of our objects, a state trajectory is a sequence of states, i.e., $\bX^k = \{ \bx^k_t \}_{t=1}^{T_k}$, where $\bx^k_t \in \cX$ for $t = 1, \ldots, T_k$.
Similarly, letting $\cW = \bbR^\contextdim$ denote our \emph{context space}, a context trajectory is a sequence of contexts, i.e., $\bW^k = \{ \bw^k_t \}_{t=1}^{T_k}$, where $\bw^k_t \in \cW$ for $t = 1, \ldots, T_k$.
Finally, letting $\cA = \bbR^\metadim$ denote the \emph{metadata space} for our objects, the metadata vector $\ba^k \in \cA$ is a static vector associated with each object.

In the ship trajectory prediction task that we consider, the state $\bx^k_t$ includes elapsed time since trajectory start in minutes, the ship's latitude, longitude, speed over ground (SOG), and course over ground (COG) at time $t$.
Then, the context $\bw^k_t$ corresponds to the significant wave height and wind speed at the same latitude and longitude as the ship, and at the same time $t$.
Finally, the metadata $\ba^k$ includes vessel hull characteristics (draft, length, and width) and cargo type. In practice, both context and metadata features may be partially unobserved; we augment each with binary missingness indicators, yielding $d_w = 3$ (wind speed, significant wave height, and a wave-height missingness indicator) and $d_a=9$ (four physical attributes and their associated indicators, plus MMSI).
An example trajectory is given in Fig.~\ref{fig:sample-tabular-data}.

\subsection{Context-Informed Trajectory Prediction}
\newcommand{\tidx}{{k^*}}

For our prediction task, we consider a fixed \emph{history length} $L$ and \emph{prediction horizon} $P$, such that $L + P \leq T_{k^*}$, where $\tidx$ is the index of our target object.
At test time, we are given:
\begin{enumerate}
    \item The historical state trajectory $\bX^\tidx_\past \defeq \{ \bx^\tidx_t \}_{t=1}^L$,
    \item The historical context trajectory $\bW^\tidx_\past \defeq \{ \bw^\tidx_t \}_{t=1}^L$, \emph{or} an indicator (e.g., $\None$) that it is unavailable, and
    \item A metadata vector $\ba^\tidx \in \cA$, \emph{or} an indicator (e.g., $\None$) that the metadata is unavailable.
\end{enumerate}
We train the model by minimizing the mean squared error (MSE) over the
full predicted state vector. For model selection and evaluation, we use
the average Haversine distance between predicted and true geographic
coordinates,
\begin{equation}
  m\!\left(Y^{k^*},\, \hat{Y}^{k^*}\right)
  = \frac{1}{P}\sum_{t=L+1}^{L+P}
    \mathrm{hav}\!\left(x_t^{k^*},\, \hat{x}_t^{k^*}\right),
\end{equation}
where $\mathrm{hav}(x, x')$ is the great-circle distance computed from
the latitude and longitude coordinates of $x$ and $x'$. We select this
metric for checkpointing because accurate geographic position prediction
is the primary operational objective; ancillary state variables (speed,
course) are of secondary concern for maritime domain awareness. We prefer the mean over the full horizon to the terminal-point metric $\text{hav}(\hat{x}^{k^*}_{L+P}, x^{k^*}_{L+P})$ because maritime applications such as collision avoidance and route de-confliction require accurate intermediate waypoint predictions throughout the horizon, not only the final position.

In our application to ship trajectory prediction, the potential unavailability of the context and metadata aligns with common real-world situations, e.g. weather data may be unavailable due to coverage gaps or the latency of weather feeds, while ship metadata may be unavailable or unreliable due to receiver outages or strategic manipulation.

\subsection{Training Regimes and Related Learning Setups}\label{subsec:DMM}

To learn a trajectory predictor, we will consider several different training regimes, with training datasets where $k \neq k^*$:
\begin{itemize}
    \item In the \emph{state-only} (S) regime, we have a training dataset consisting only of state trajectories $\bX^k$.
    \item In the \emph{state+metadata} (S+M) regime, we have a training dataset of tuples $(\bX^k, \ba^k)$.
    \item Similarly, in the \emph{state+context} (S+C) regime, we have a training dataset of tuples $(\bX^k, \bW^k)$.
    \item Finally, in the \emph{state+context+metadata} (S+C+M) regime, we have a training dataset of tuples $(\bX^k, \bW^k, \ba^k)$.
\end{itemize}
Notably, the data modalities available during training may differ from those available during inference time, e.g., if we consider predicting when the context trajectory and metadata are unavailable (i.e., $\bW^\tidx_\past = \None$ and $\ba^\tidx = \None$) after training in the S+C+M regime.
We refer to this possibility as the \emph{deployment modality mismatch (DMM)} problem, which is a key design consideration for our proposed architecture and training strategy.

The DMM problem is closely related to two existing learning setups.
First, the DMM setup generalizes \emph{learning using privileged information (LUPI)}~\citep{vapnik2009new,vapnik2015learning,lopez2015unifying}, where the additional modalities are guaranteed to be absent during inference time (or must simply be ``thrown out" if they are present).
Second, the DMM setup is an important special case of the more general idea of \emph{learning with side information (LWSI)}~\citep{jonschkowski2015patterns,adel2023general}, which considers a wider variety of inputs, e.g. facts from a knowledge base.
By concentrating on the DMM setup, we can benefit from the additional modalities when they are available at inference time (unlike LUPI), while focusing on the specific problem structure (unlike LWSI in its more general form).

%% file: content/methods.tex
\section{Methods}

We propose the Conditional InformerTP, an encoder-decoder Transformer architecture designed to efficiently fuse heterogeneous maritime data. The architecture consists of three key components: a Multimodal Embedding Layer, a Conditional Encoder, and a Conditional Decoder.

\subsection{Base Architecture}

Our proposed model builds upon the Informer architecture~\citep{zhou2021informer}, and its extension Informer-TP~\citep{xiong2024informer}, a specialized Transformer designed for long-term ship trajectory prediction. We leverage its ProbSparse Attention to reduce complexity to $O(L \log L)$ , and Self-Attention Distilling (1-D convolution and MaxPool) to downsample redundant features in long sequences. We also adopt Informer-TP’s domain-specific embedding layers for kinematic states.

\subsection{Motivation}

Unlike standard models that treat vessel motion as an isolated system, maritime navigation is physically modulated by extrinsic forces like wind and waves. We employ an encoder-decoder structure to model the conditional distribution $P(Y|X,W)$ rather than a symmetric joint distribution $P(X,W)$. This ensures a directional information flow where the Encoder acts as a Context Modeler (aggregating $X$ and $W$) and the Decoder acts as a State Generator (querying the context via cross-attention).

\subsection{Conditional InformerTP}

\subsubsection{Embedding}
\label{sec:embedding}

To handle the heterogeneous nature of the inputs, we employ three distinct embedding streams.
Each temporal stream (state and context) produces a token embedding that is the sum of three components:
a \emph{value projection}, a \emph{positional encoding}, and a \emph{temporal embedding}.

\paragraph{Value Projection.}
The normalized state vector $\bx^k_t \in \cX$ (where $d_x = dim(\cX) = 5$: elapsed time, latitude, longitude, SOG, COG)
is mapped to a $d_X$-dimensional representation by a learnable linear projection (bias-free):
\begin{equation}
  u_t = \bx^k_t \, W_{\text{val}}, \quad W_{\text{val}} \in \mathbb{R}^{d_x \times d_X}.
\end{equation}
The context vector $\bw^k_t \in \cW$ 
undergoes an analogous projection to $\mathbb{R}^{d_W}$.

\paragraph{Positional Encoding.}
To preserve sequence ordering---particularly important given that ProbSparse attention
selects a subset of query positions---we add a fixed sinusoidal positional encoding
$\mathrm{PE}_t \in \mathbb{R}^{d_X}$ following~\citet{vaswani2017attention}.

\paragraph{Temporal Embedding.}
To capture absolute temporal patterns (e.g., time-of-day effects on shipping traffic,
seasonal weather variation), we extract five calendar features from each timestamp:
minute, hour, ISO weekday, month, and year.
Each is normalized to approximately $[-0.5,\, 0.5]$ and the five scalars are
jointly projected to $d_X$ via a pointwise convolution
($\texttt{Conv1d}$ with kernel size~1):
\begin{equation}
  \mathrm{TE}_t = \texttt{Conv1d}_{5 \to d_X}\!\bigl([\tilde{c}_1,\, \tilde{c}_2,\, \tilde{c}_3,\, \tilde{c}_4,\, \tilde{c}_5]_t\bigr),
\end{equation}
where $\tilde{c}_i$ denotes the normalized $i$-th calendar scalar.

\paragraph{State and Context Embeddings.}
The final embedding for the state stream is the summation of the three components:
\begin{equation}
  H_X^{(t)} = u_t + \mathrm{PE}_t + \mathrm{TE}_t.
\end{equation}
The context stream follows an identical embedding procedure
(with its own projection $W_{\text{val}}^{(W)} \in \mathbb{R}^{d_w \times d_W}$
and temporal embedding $\texttt{Conv1d}_{5 \to d_W}$),
sharing the same calendar features to ensure temporal alignment between
the vessel state and its environmental conditions.

\paragraph{Metadata Embedding.}
Static vessel attributes (draft, length, width, cargo type, associated
missing indicators, and MMSI) are min--max normalized and concatenated into a single vector
$\ba^k \in \mathbb{R}^{d_a}$, which is then projected through a linear layer:
\begin{equation}
  H_a = \ba^k \, W_{\text{meta}}, \quad W_{\text{meta}} \in \mathbb{R}^{d_a \times d_M}.
\end{equation}
Since $H_a$ is static (identical across time steps for a given trajectory),
it is broadcast to the temporal dimension during conditioning
(Section~\ref{sec:conditional_attention}).

\subsubsection{Conditional Attention}
\label{sec:conditional_attention}

The Conditional Attention layer enables the state stream to selectively
incorporate information from auxiliary modalities while preserving
modality-specific structure. It consists of three parallel branches---state
self-attention, context cross-attention, and metadata conditioning---whose
outputs are summed and passed through a shared output projection.

Let $H_X \in \mathbb{R}^{B \times L \times d_X}$ denote the state
embedding, $H_W \in \mathbb{R}^{B \times L \times d_W}$ the context
embedding, and $H_a \in \mathbb{R}^{B \times d_M}$ the metadata
embedding. We write $n_h$ for the number of attention heads and define
the per-head dimensions $d_k = d_v = d_X / n_h$.

\paragraph{State Self-Attention (ProbSparse).}
Queries, keys, and values are obtained from independent linear projections
of the state embedding:
\begin{equation}
  Q_X = H_X W_X^Q,\quad K_X = H_X W_X^K,\quad V_X = H_X W_X^V,
\end{equation}
where $W_X^Q, W_X^K \in \mathbb{R}^{d_X \times d_k n_h}$ and
$W_X^V \in \mathbb{R}^{d_X \times d_v n_h}$.
The self-attention output is computed via ProbSparse attention~\citep{zhou2021informer}:
\begin{equation}
  O_X = \mathrm{ProbSparseAttn}(Q_X,\, K_X,\, V_X),
\end{equation}
which selects the top-$u$ dominant queries
($u = c\,\lceil \ln L \rceil$ for sparsity factor $c$) and computes
scaled dot-product attention only at those positions, reducing
complexity from $O(L^2)$ to $O(L \log L)$.

\paragraph{Context Cross-Attention (Full).}
To condition the state on the environmental context, we compute a
cross-attention in which queries are derived from the state embedding
and keys/values from the context embedding.
Crucially, this branch uses a \emph{separate} query projection from
the self-attention branch, allowing the model to learn distinct
``questions to ask'' of itself versus of the environment:
\begin{align}
  Q_{X \to W} = H_X\, W_{X \to W}^Q, 
  K_W = H_W\, W_W^K, V_W = H_W\, W_W^V, \label{eq:cross_qkv}
\end{align}
where $W_{X \to W}^Q \in \mathbb{R}^{d_X \times d_k n_h}$,
$W_W^K \in \mathbb{R}^{d_W \times d_k n_h}$, and
$W_W^V \in \mathbb{R}^{d_W \times d_v n_h}$.
Note that the query and key projections map from different input
dimensions ($d_X$ and $d_W$, respectively) into the common
$d_k$-dimensional head space, enabling dot-product compatibility
scores to be computed across modalities.
The cross-attention output is:
\begin{equation}
  O_W = \mathrm{softmax}\!\left(\frac{Q_{X \to W}\, K_W^\top}{\sqrt{d_k}}\right) V_W.
  \label{eq:cross_attn}
\end{equation}
This directional design ensures that the state queries the context
but not vice versa, encoding the physical asymmetry that the
environment modulates---but is not generated by---vessel dynamics.

\paragraph{Metadata Conditioning (Project and Broadcast).}
Static vessel attributes $a^k$ (e.g., draft, length, cargo type) are projected to $H_a \in \mathbb{R}^{B \times d_M}$. Since $H_a$ is constant across all time steps $L$, applying standard dot-product attention over this single key-value pair results in a softmax weight of exactly $1$ at every query position . Consequently, the attention mechanism reduces to a linear projection followed by a temporal broadcast: $O_a = 1_L \otimes (H_a W_a^V)$ . This formulation avoids redundant query-key projections while maintaining mathematical equivalence to full cross-attention, allowing us to bypass the mechanism entirely:
\begin{equation}
  O_a = \mathbf{1}_L \otimes \bigl(H_a\, W_a^V\bigr),
  \label{eq:meta_broadcast}
\end{equation}
where $W_a^V \in \mathbb{R}^{d_M \times d_v n_h}$ and
$\mathbf{1}_L \otimes$ denotes broadcasting the projected vector
to all $L$ time steps.
The metadata thus acts as a \emph{global additive bias} that
uniformly conditions every position in the trajectory.%
\footnote{%
  Some of our experiments retain the full query-key-value
  formulation for metadata. Since the softmax over a single
  key collapses to unity, the two implementations are
  mathematically equivalent; we present the simplified form
  here for clarity.%
}

\paragraph{Summation.}
The final output of the Conditional Attention layer is the element-wise
sum of the three branches, followed by an output projection:
\begin{equation}
  O_{\text{final}} = \bigl(O_X + O_W + O_a\bigr)\, W^O,
  \label{eq:aggregation}
\end{equation}
where $W^O \in \mathbb{R}^{d_v n_h \times d_X}$.
This additive fusion ensures that when an auxiliary modality is
masked---either via Modality Masking during training
(Section~\ref{sec:modality_dropout}) or due to unavailability at
inference---its contribution is zeroed out without disrupting the
tensor shapes of the primary state stream.

\subsubsection{Conditional Encoder}

Figure \ref{fig:encoder} illustrates our encoder architecture. The encoder consists of a stack of Conditional Encoder Layers. In each layer, the vessel trajectory $\mathbf{H}_X$ attends to itself via self-attention while simultaneously attending to the context $\mathbf{H}_W$ and metadata $\mathbf{H}_a$. The output of the $i$-th layer is defined as:

\[\mathbf{H}_X^{(i+1)} = \text{LayerNorm}\left( \mathbf{H}_X^{(i)} + \text{CondAttn}(\mathbf{H}_X^{(i)}, \mathbf{H}_W, \mathbf{H}_a) \right)\]

Following the standard Informer design, we also employ distilling operations (Conv1d + MaxPool) between layers to reduce the sequence length and highlight dominant features.

\subsubsection{Conditional Decoder}
Figure \ref{fig:decoder} illustrates our decoder architecture.
The decoder aims to generate the future trajectory $\mathbf{H}_{pred}$. It utilizes a standard masked self-attention layer to model the temporal dependencies of the prediction sequence itself. However, for the cross-attention step---where the decoder attends to the encoder's output---we replace the standard mechanism with a Conditional Cross-Attention Layer. In this layer, the query comes from the decoder's current state (the generated future trajectory so far). This query attends to three separate sets of keys/values provided by the encoder: the encoded past trajectory states, the encoded context, and the encoded metadata. This architecture ensures that the generation of future positions is directly conditioned on both the historical motion pattern and the prevailing weather conditions.

\input{figures/arch_img}

\subsection{Training Strategy}

\subsubsection{Causal Masking}
To ensure the validity of prediction, we apply a regular Triangular Causal Mask in the decoder's self-attention layers. This mask $M \in \{0, -\infty\}^{P \times P}$ sets the attention score to $-\infty$ for all upper-triangular elements, preventing the model from attending to future positions $j > t$ when predicting the state at time $t$.

\subsubsection{Modality Masking}
A model trained on complete data triplets $\{\mathbf{X}, \mathbf{W}, \mathbf{a}\}$ often learns to over-rely on the auxiliary modalities, leading to catastrophic performance degradation when they are missing at inference time. To enforce robustness, we introduce a Modality Masking strategy during training. For each training batch, we independently mask the context $\mathbf{W}$ and the metadata $\mathbf{a}$ with probabilities $p_{W}$ and $p_{a}$ respectively.

Formally, let $b_{W}, b_{a} \sim \text{Bernoulli}(1 - p)$ be binary indicator variables where $1$ indicates presence and $0$ indicates absence. The input to the model during a training step becomes:$$\mathbf{W}' = b_{W} \cdot \mathbf{W}, \quad \mathbf{a}' = b_{a} \cdot \mathbf{a}$$In our implementation, masking is applied batch-wise: when a modality is dropped, its corresponding tensor is replaced with $\None$. Due to the summation computation of our Conditional Attention mechanism, this zeroes out the contribution of that branch.

%% file: figures/arch_img.tex
\begin{figure*}[t]
    \centering
    
    \begin{subfigure}{0.45\textwidth}
        \centering
        \includegraphics[width=\linewidth]{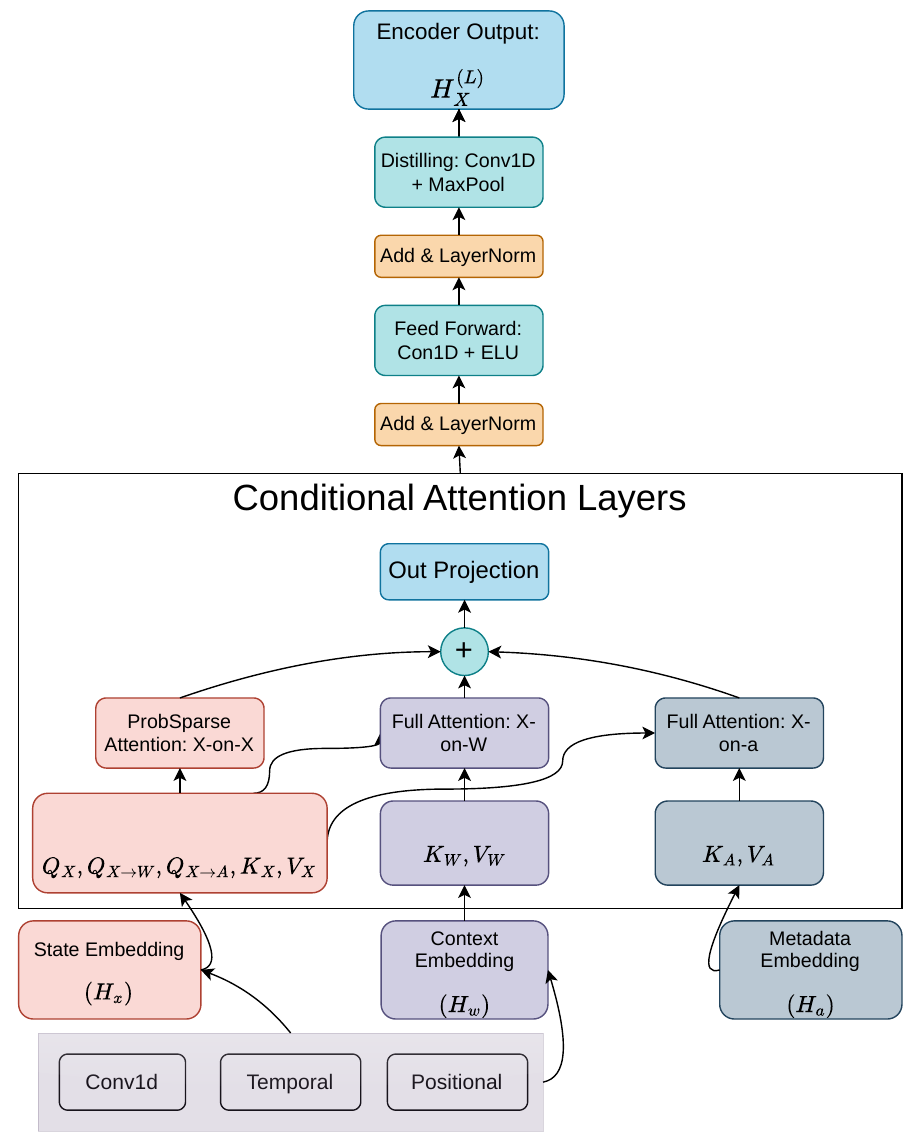}
        \caption{Encoder}
        \label{fig:encoder}
    \end{subfigure}
    \hfill
    \begin{subfigure}{0.45\textwidth}
        \centering
        \includegraphics[width=\linewidth]{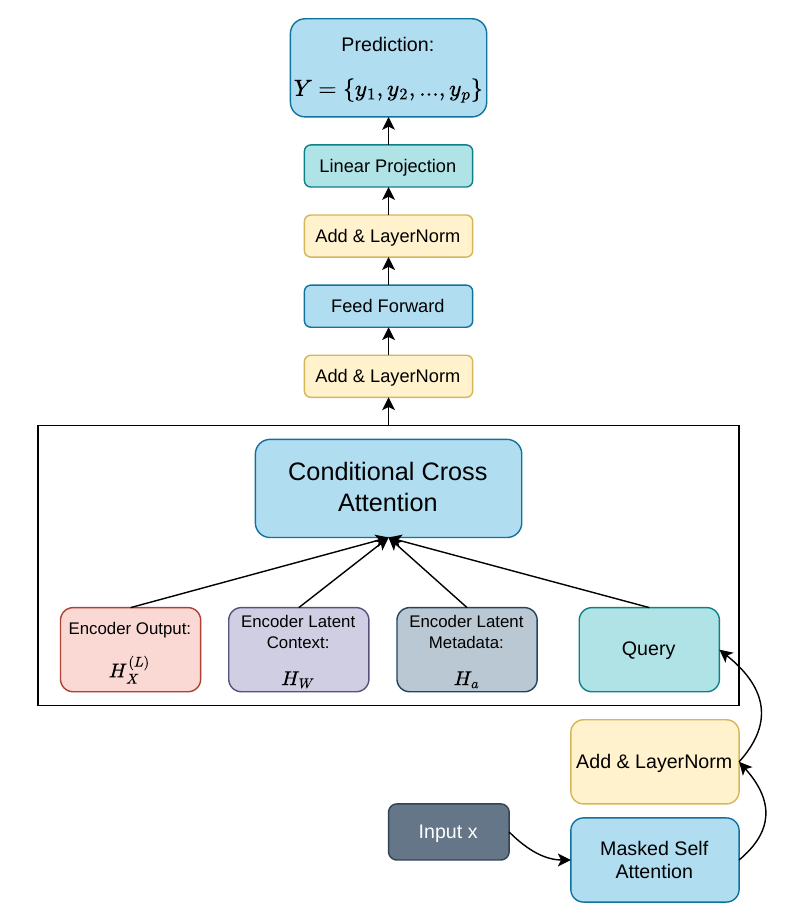}
        \caption{Decoder}
        \label{fig:decoder}
    \end{subfigure}
    
    \caption{\textbf{Architecture of the Conditional Informer.}
  (a) Encoder with state self-attention, context cross-attention, and metadata bias. (b) Decoder with masked self-attention and conditional cross-attention over encoder representations.}
  \label{fig:architecture}
\end{figure*}

%% file: content/experiments.tex
\section{Experiments}
In this section, we evaluate the proposed Conditional Informer against varied baselines to validate the effectiveness of incorporating environmental context and the robustness of our conditional architecture.

\subsection{Datasets and Preprocessing}
\begin{itemize}
\item \textbf{Vessel Trajectories (AIS):} Raw Automatic
  Identification System (AIS) data was sourced from the Marine
  Cadastre AIS vessel traffic archive~\citep{marinecadastre}, which provides over 100,000
  open-source trajectories spanning 2009--2025 across diverse vessel
  types. Our study focuses on the period from June 2023 to September
  2023 within the Gulf of Mexico region (Lat: 23--27$^\circ$N,
  Lon: 84--77$^\circ$W). We focus on cargo vessels (IMO type 70) to maintain a homogeneous vessel population and to remain methodologically consistent with Informer-TP~\citep{xiong2024informer}, which evaluates on the same vessel class and region. The conditional architecture is vessel-type agnostic; extension to heterogeneous fleets is a natural direction for future work (Section~\ref{sec:discussion}).

  Following the pipeline of~\citet{xiong2024informer},
  data cleaning consisted of four steps: (1) DBSCAN outlier removal with $\varepsilon$ estimated as the 95th percentile of the 5-nearest-neighbor distance distribution; (2) gap-based segmentation, splitting trajectories at intervals exceeding 60 minutes; (3) linear interpolation (for continuous variables) and forward/backward filling (for categorical attributes) to bridge minor dropouts under $1$ minute; and (4) uniform resampling to $5$-minute intervals.

\item \textbf{Environmental Context (ERA5):} To enrich the kinematic
  trajectories, we integrated environmental features from the ERA5
  reanalysis dataset~\citep{hersbach2020era5}, obtained from the Climate Data Store \citep{era5_cds}.
  The dataset provides a global latitude-longitude grid
  of atmospheric and oceanic state variables at hourly resolution.
  We extracted two features: 10-meter wind speed (m/s) and significant
  wave height (m). Given the resolution mismatch between the
  high-frequency AIS stream and the hourly, discrete ERA5 grid
  ($0.25^\circ$), we employed a nearest-neighbor lookup strategy
  in both space and time. Each AIS point was assigned environmental
  variables from the spatially nearest ERA5 grid coordinate and the
  preceding full hour. Missing significant wave height values (undefined
  over certain regions and time periods) were zero-imputed after
  normalization.

\item \textbf{Static Metadata:} We included vessel
  dimension features---draft, length, and width---along with the
  cargo type attribute. These physical characteristics capture the
  vessel's inertia and hydrodynamic profile, which directly influence
  its responsiveness to environmental forces. Missing dimension values
  were zero-imputed with corresponding binary missing indicators
  appended to the feature vector.
\end{itemize}

Trajectory segments with non-recoverable missing data or fewer than 48 resampled points (4 hours) were discarded, ensuring all tracks comfortably exceeded the $L+P=36$ step model requirement. The remaining 1,571 valid trajectory segments were split strictly at the trajectory level (80/10/10) to prevent data leakage across sets, yielding 1,256 training, 157 validation, and 158 test tracks. Finally, a sliding window with a history length of $L=24$ (2 hours) and prediction horizon $P=12$ (1 hour) was applied to these segments. This produced a final dataset of 123,771 samples (98,944 training / 12,785 validation / 12,042 test).
\input{tables/sample_table}

\subsection{Baselines and Experimental Setup}
We compare the following configurations:
\begin{enumerate}
    \item \textbf{InformerTP (State-Only):} The baseline Informer model that utilizes only kinematic features (Latitude, Longitude, SOG, COG).
    \item \textbf{InformerTP (Concatenated):} A fusion baseline where environmental context and metadata embeddings are concatenated directly with the state embeddings at the input level, treating context simply as additional feature channels.
    \item \textbf{Conditional Informer (Ours):} Our proposed model using Conditional Attention to fuse state, context, and metadata via separate pathways.
\end{enumerate}

All models were trained with a model dimension $d_{model}=512$, $n_{heads}=8$, and a 2-layer encoder/1-layer decoder architecture. We utilized the Adam optimizer with initial learning rate of $5\times10^{-6}$ and betas of $(0.9, 0.98)$. To prevent overreliance on other modalities, we applied a masking rate of 0.4 for context and 0.3 for metadata---training employed mixed-precision (BF16) to optimize memory usage on 2x NVIDIA A6000 GPUs. The model was trained for a maximum of 100 epochs with an early stopping patience of 20 epochs, monitoring the validation Haversine distance to save the best-performing checkpoint. Training took about 2 hours per model, and inference took approximately $7.5$ ms for a single-vessel forward pass.

%% file: tables/sample_table.tex
% \begin{table}[htbp]
% \caption{Results}
% \centering
% \resizebox{\linewidth}{!}{%
% \begin{tabular}{|c|c|c|c|}
% \hline
% \textbf{Ablation}&\multicolumn{3}{|c|}{\textbf{Input Data}} \\
% \cline{2-4} 
% \textbf{Results} & \textbf{\textit{state only}}& \textbf{\textit{state+context}}& \textbf{\textit{state+context+metadata}} \\
% \hline
% MSE& More table copy$^{\mathrm{a}}$& &  \\
% \hline
% \multicolumn{4}{l}{$^{\mathrm{a}}$Sample of a Table footnote.}
% \end{tabular}%
% }
% \label{tab1}
% \end{table}

\begin{table}[htbp]
\centering
\caption{\textbf{Main Results}: Haversine distance (km) and MSE ($\times 10^{-3}$) comparison. Each cell reports mean $\pm$ std across 5 random seeds. \textbf{Bold} marks the best result per column. Metadata ($M$) shows limited independent impact due to vessel population homogeneity, while Context ($C$) drives a $\sim$15.4\% accuracy gain over the baseline.}
\label{tab:main_results}

\setlength{\tabcolsep}{3pt}
% \resizebox constrains the wide 9-column table to fit the single-column width
\resizebox{\linewidth}{!}{%
\begin{tabular}{|l||c|c|c|c||c|c|c|c|}
\hline
\textbf{Model} & \multicolumn{4}{c||}{\textbf{Haversine (km) [Test Time Data]}} & \multicolumn{4}{c|}{\textbf{MSE ($\times 10^{-3}$) [Test Time Data]}} \\
\cline{2-9}
\textbf{(Training Time Data)} & \textbf{\textit{State Only}} & \textbf{\textit{S+C}} & \textbf{\textit{S+M}} & \textbf{\textit{S+C+M}} & \textbf{\textit{State Only}} & \textbf{\textit{S+C}} & \textbf{\textit{S+M}} & \textbf{\textit{S+C+M}} \\
\hline
\hline
\textbf{InformerTP-(S)} & 6.03 $\pm$ 0.22 & N/A & N/A & N/A & 1.46 $\pm$ 0.21 & N/A & N/A & N/A \\
\hline
\textbf{Concat-(S+C+M)} & 9.28 $\pm$ 1.26 & 9.20 $\pm$ 1.19 & 6.47 $\pm$ 0.56 & 6.12 $\pm$ 0.33 & 1.93 $\pm$ 0.20 & 1.81 $\pm$ 0.16 & 1.55 $\pm$ 0.14 & 1.50 $\pm$ 0.17 \\
\hline
\textbf{Conditional-(S)} & \textbf{5.97 $\pm$ 0.44} & N/A & N/A & N/A & \textbf{1.41 $\pm$ 0.17} & N/A & N/A & N/A \\
\hline
\textbf{Conditional-(S+M)} & 6.04 $\pm$ 0.52 & N/A & \textbf{6.02 $\pm$ 0.58} & N/A & 1.45 $\pm$ 0.19 & N/A & \textbf{1.45 $\pm$ 0.19} & N/A \\
\hline
\textbf{Conditional-(S+C)} & 8.89 $\pm$ 1.16 & \textbf{5.10 $\pm$ 0.33} & N/A & N/A & 1.52 $\pm$ 0.22 & \textbf{1.40 $\pm$ 0.18} & N/A & N/A \\
\hline
\textbf{Conditional-(S+C+M)} & 8.76 $\pm$ 1.64 & 5.13 $\pm$ 0.31 & 8.75 $\pm$ 1.67 & \textbf{5.10 $\pm$ 0.36} & 1.50 $\pm$ 0.22 & 1.41 $\pm$ 0.17 & 1.53 $\pm$ 0.22 & \textbf{1.40 $\pm$ 0.16} \\
\hline
\hline
\textbf{No masking-(S+C+M)} & 67.05 $\pm$ 12.84 & 7.04 $\pm$ 0.44 & 66.19 $\pm$ 13.36 & 6.44 $\pm$ 0.49 & 8.40 $\pm$ 2.06 & 1.50 $\pm$ 0.21 & 8.49 $\pm$ 2.09 & 1.51 $\pm$ 0.21 \\
\hline
\end{tabular}%
}
\end{table}

%% file: content/results.tex
\section{Results and Analysis}
\label{sec:ablation}
\subsection{Comparison to Baselines}

Table~\ref{tab:main_results} summarizes the performance of all model
configurations. We compare the proposed Conditional Informer against
the InformerTP baseline (state-only) and InformerTP-Concat
(concatenated multimodal fusion). All results report mean and
standard deviation over 5 random seeds.

When all modalities are available at test time, the Conditional
Informer consistently outperforms both baselines. The best overall
result is achieved by the Conditional (S+C+M) model with full inputs,
yielding a Haversine error of $5.10$ km---a $15.4\%$ reduction compared
to the InformerTP baseline ($6.03$ km) and a $16.7\%$ improvement over
the Concat model's full-modality result ($6.12$ km). These gains confirm
that our conditional attention mechanism more effectively leverages
environmental context than na\"ive concatenation, by enforcing the
causal direction from weather conditions to vessel dynamics rather
than treating all modalities symmetrically.

The Conditional (S) model, which uses the same architecture but with
auxiliary branches disabled, achieves $5.97$ km---on par with the
InformerTP baseline ($6.03$ km) given overlapping standard deviations.
This parity is expected by construction: when auxiliary modalities
are absent, the architecture naturally reduces to the baseline's computation, confirming that the additional branches introduce no interference.
\subsection{Effect of Environmental Context}
Environmental context emerges as the primary driver of prediction
improvement. The Conditional (S+C) model achieves $5.10$ km when
context is provided, compared to 8.89~km without---a $42.6\%$
reduction in error attributable solely to wind and wave information.
Similarly, the Conditional (S+C+M) model improves from $8.76$ km
(state-only) to $5.10$ km (full inputs) when context is available.
This pattern is consistent with the physical reality that environmental
forces directly modulate vessel ground speed and heading, providing
predictive signal that cannot be recovered from kinematic history alone.

In contrast, the Concat baseline derives comparatively less benefit
from context: its full-modality result ($6.12$ km) improves only
modestly over state+metadata ($6.47$ km), and its state+context
result ($9.20$ km) barely improves over state-only ($9.28$ km). This
suggests that the concatenation approach struggles to disentangle
the causal influence of weather from the kinematic signal, whereas
the conditional architecture's separated attention pathways enable
more effective use of environmental data.

We additionally evaluated a shorter history length of $L=P=12$. In this setting, context provides negligible benefit: Conditional-(S+C) achieves $5.97$ km with context versus $5.99$ km without, compared to the $42.6\%$ gain at $L=24$. We attribute this to insufficient temporal signal at $L=12$ for the model to identify weather-driven deviations from the nominal kinematic trajectory, corroborating that a two-hour context window is necessary to observe environmentally-modulated behavioral patterns.

\subsection{Effect of Metadata}

Vessel metadata shows limited
independent impact on prediction accuracy. The Conditional (S+M) model
achieves $6.02$ km with metadata and $6.04$ km without---a negligible
difference that falls within the standard deviation. Similarly, the
full S+C+M model's performance with all modalities ($5.10$ km) is
essentially identical to the S+C model with context alone
($5.10$ km), suggesting that metadata provides minimal additional
signal beyond what context already captures.

\subsection{Modality Masking}\label{sec:modality_dropout}

The ``Without Masking" ablation demonstrates the necessity of Modality Masking. Without it, the Conditional (S+C+M) model catastrophically fails when environmental data is missing at inference, with state-only error reaching $67.05$ km—an order of magnitude worse than the masked-trained model $(8.76$ km) and the kinematic baseline ($6.03$ km). This confirms the unmasked model treats weather as a non-optional shortcut rather than learning a robust kinematic representation. Furthermore, masking acts as a regularizer that improves full-modality performance: the masked S+C+M model achieves $5.10$ km compared to the unmasked variant's $6.44$ km (a $20.8\%$ improvement).

When context is withheld at test time, the context-trained models naturally degrade, validating that they actively utilize wind and wave vectors as physical drivers of trajectory adjustment. The state-only fallback errors increase to $8.89$ km for Conditional (S+C) and $8.76$ km for (S+C+M). Although slightly worse than the state-only baseline ($6.03$ km), this represents a manageable degradation compared to the unmasked model's failure ($67.05$ km).

%% file: content/discussion.tex
\section{Discussion}\label{sec:discussion}

%It is important to note that the proposed Conditional Attention
%mechanism is not architecturally bound to the Informer framework.
%Because it relies on a fundamental query-key-value separation---where
%the state trajectory queries the auxiliary contexts---this module
%can be integrated into any attention-based architecture, including
%standard Transformers, Performers, or newer State Space Models.~\cite{gu2022efficiently}

%Several directions remain for future investigation. Enriching the
%environmental context with ocean currents and bathymetry (water
%depth) data would provide stronger physical constraints. Expanding
%the vessel population to include multiple vessel types would
%test whether the metadata branch contributes more substantially
%under greater fleet heterogeneity. Investigating adaptive dropout
%schedules or auxiliary loss terms to further close the gap between
%context-trained models' fallback performance and the state-only
%baseline remains an important practical challenge. Finally, extending
%the framework to incorporate additional exogenous signals---such as
%port schedules or geopolitical routing
%constraints---could further improve prediction in operationally
%complex scenarios.
It is important to note that the proposed Conditional Attention mechanism is not architecturally bound to the Informer framework. Because it relies on a fundamental query-key-value separation --- where the state trajectory queries the auxiliary contexts --- this module can be integrated into any attention-based architecture, including standard Transformers, Performers, or State Space Models~\citet{gu2022efficiently}.

\textbf{Limitations.} Although Modality Masking substantially mitigates feature collapse, it does not fully eliminate the DMM problem (Section~\ref{subsec:DMM}): when context is absent at test time, Conditional-(S+C+M) reaches $8.76$ km versus $6.03$ km for the kinematic baseline. This residual degradation suggests the current masking still permits some shortcut reliance on auxiliary data; future work on adaptive masking schedules or auxiliary contrastive objectives may close this gap further.

The present evaluation focuses on open-water transit trajectories typical of cargo vessels. Whether the model accurately captures non-trivial behaviors—sudden course changes, speed modulations in heavy weather, or port approach maneuvers—remains an open question that requires targeted qualitative analysis.

Finally, the study is limited to cargo vessels (type $70$). Expanding to a heterogeneous fleet—tankers, container ships, fishing vessels—would test whether the metadata branch contributes more substantially under greater variation in hull characteristics, and is a priority for future work.

Several additional directions remain. Enriching the environmental context with ocean currents and bathymetry would provide stronger physical constraints. Extending the framework to incorporate port schedules or geopolitical routing constraints could further improve prediction in operationally complex scenarios.\\

\textbf{Data Availability Statement.}
The AIS data were obtained from \url{https://marinecadastre.gov/ais/} and are publicly available; no restrictions on use apply beyond appropriate attribution.
The ERA5 reanalysis data were obtained from the Climate Data Store (\url{https://cds.climate.copernicus.eu/datasets/reanalysis-era5-single-levels}), provided by the Copernicus Climate Change Service.
The data are distributed under the Creative Commons Attribution 4.0 International (CC-BY 4.0) license; no restrictions on use apply by appropriate attribution.